%% file: main.tex
\documentclass[lettersize,journal]{IEEEtran}
\input{prefix}
\begin{document}
\pagenumbering{arabic}

\title{Combining Federated Learning and \\Control: A Survey*}

\author{Jakob Weber$^{1}$, Markus Gurtner$^{1}$, Amadeus Lobe$^{1}$, Adrian Trachte$^{2}$, and Andreas Kugi$^{3}$% <-this % stops a space
\thanks{$^{1}$ Jakob Weber, Markus Gurtner, and Amadeus Lobe are with the Center for Vision, Automation \& Control, AIT Austrian Institute of Technology GmbH, Vienna, Austria 
                {\tt\small jakob.weber@ait.ac.at}}
\thanks{$^{2}$ Adrian Trachte is with the Robert Bosch GmbH, Renningen, Germany}
\thanks{$^{3}$ Andreas Kugi is with the Automation and Control Institute, TU Wien, Austria and the AIT Austrian Institute of Technology GmbH, Vienna, Austria}
}

\maketitle
\thispagestyle{plain}  % force page number

\begin{abstract}
This survey provides an overview of combining Federated Learning (FL) and control to enhance adaptability, scalability, generalization, and privacy in (nonlinear) control applications. 
Traditional control methods rely on controller design models, but real-world scenarios often require online model retuning or learning. 
FL offers a distributed approach to model training, enabling collaborative learning across distributed devices while preserving data privacy. 
By keeping data localized, FL mitigates concerns regarding privacy and security while reducing network bandwidth requirements for communication. 
This survey summarizes the state-of-the-art concepts and ideas of combining FL and control.
The methodical benefits are further discussed, culminating in a detailed overview of expected applications, from dynamical system modeling over controller design, focusing on adaptive control, to knowledge transfer in multi-agent decision-making systems.
\end{abstract}


\section{Introduction}\label{sec:Intro}

With the growing importance of data-driven models within control systems, there is an increasing emphasis on integrating learning-based models directly into the control loop.
This integration enhances adaptability and allows for broader generalizability across diverse and possibly unseen operational scenarios. 
Nowadays, an increasing number of control system hardware options include integrated connectivity solutions, exemplified by~\cite{BoschBodas} or~\cite{Bechhoff}, thereby creating opportunities, e.g., to integrate cloud-based solutions to enhance system performance, resilience, and adaptability.
While centralized approaches leveraging connected Internet-of-Things (IoT) devices and cloud computing infrastructure present viable options, bandwidth limitations and data privacy challenges are still present when transmitting raw data, see, e.g.,~\cite{sun2020privacy_in_ML} discussing privacy in the context of machine-learning and 6G communication, and~\cite{ma_2023_trusted_AI_in_multiagent_systems} for an in-depth discussion of privacy and security for distributed machine-learning techniques.
Furthermore,~\cite{park2021_comm-eff_distr_learning_over_networks} offers an excellent overview of communication-efficient and distributed learning. 
In this context, Federated Learning (FL) offers a compelling solution.

Federated Learning enables collaborative model training across distributed devices while preserving sensitive data, thereby addressing the challenges of communication efficiency and privacy preservation.
The core concept involves computing model updates locally on individual devices, securely aggregating these updates, and globally computing a combined model. 
This paradigm shift from traditional centralized to decentralized training addresses data privacy and security by keeping the raw data localized, thus ensuring minimal exposure risk. 
Beyond regulatory considerations, see~\cite{woisetschläger2024federated} for a discussion of FL under the European Union Artificial Intelligence Act~\cite{EU_2023_AI-act}, FL offers a technical advantage in reducing the required network bandwidth through a more efficient information transfer.

Federated Learning has already gained substantial attention, particularly in industries and domains characterized by abundant sensitive data with high data privacy and communication efficiency demands, such as healthcare~\cite{nvidia2019clara, chen2023networking, zhou2021_DL_in_medical_imaging}, finance and economics~\cite{nvidia2022finance, long2020_federated_learning_open_banking, wang2023_FL_collaborative_price_prediction_flexible_market, liu2023_efficient_FL_financial_applications}, manufacturing~\cite{deng2023_FL_collaborative_manufacturing, savazzi2021opportunities, hegiste_2022_FL_in_manufacturing}, and IoT applications~\cite{nguyen2021federated_IoT_survey, tallat2023industry50_FL, liu2023_enabling_AIoT_survey, brecko2022_FL_edge_computing_survey, gu2023_AI_Cloud_edge_terminal_survey, shi2020communication_efficient_edge_AI}. 
The distributed nature of FL is advantageous in scenarios featuring large datasets or devices spread across diverse spatial locations, see e.g., \cite{kairouz2021advances} and~\cite{nokleby2020_dist_processing_of_data_streams}.

\begin{figure*}[t]
  \centering
    \includegraphics[width=\textwidth]{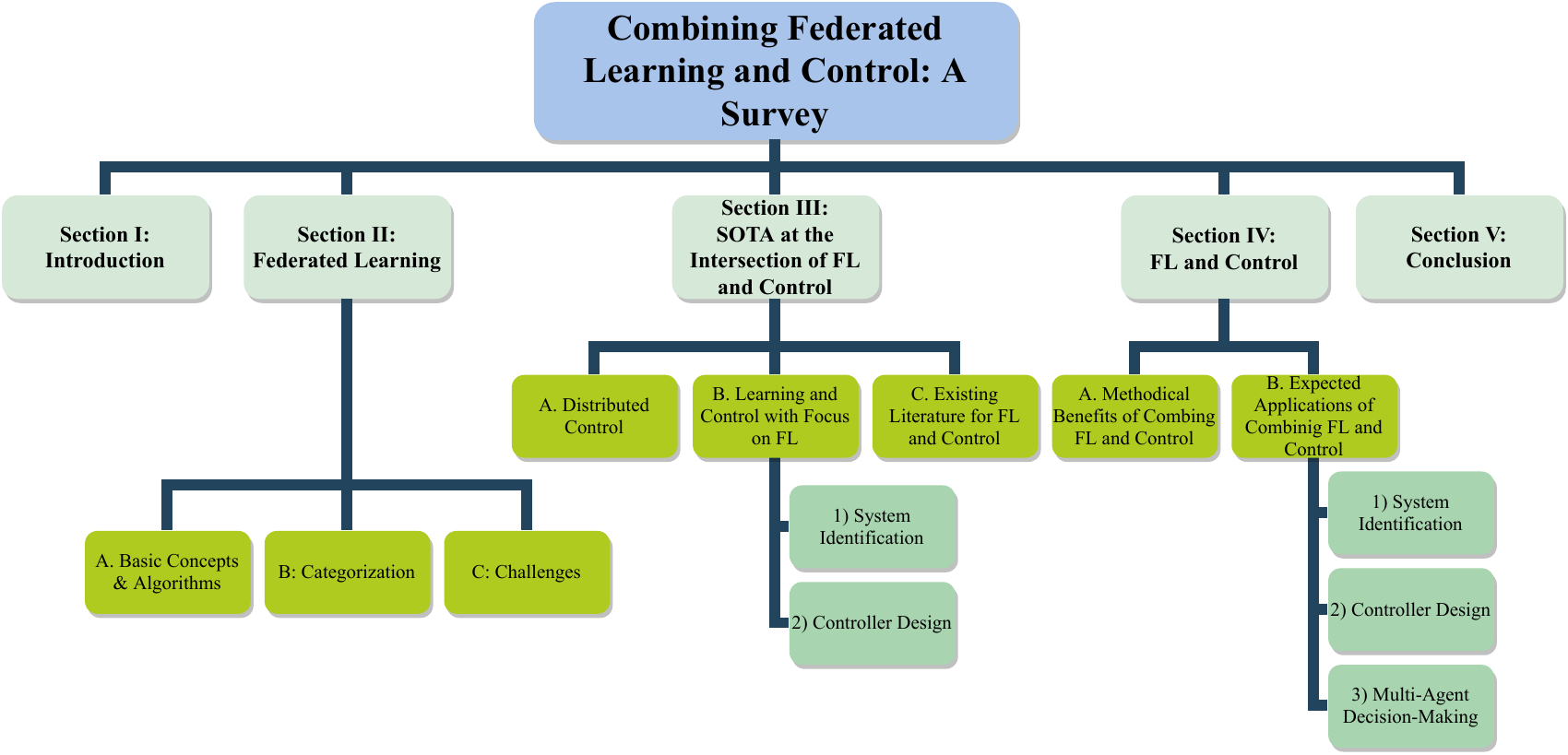}
  \caption{The structure of this paper.}
  \label{fig:organigram}
\end{figure*}

This survey aims to offer a comprehensive overview of the potential of FL for control problems, focusing on enhancing adaptability, scalability, resilience, and privacy in (nonlinear) control applications. 
Our contribution is to provide answers to the following research questions:
\begin{enumerate} [label=Q\arabic*)]
    \item What is the current state-of-the-art at the intersection of FL and control?
    \item What are the anticipated benefits of combining FL and control?
\end{enumerate}
The paper is structured as follows:
Section~\ref{sec:FL} introduces FL, presenting its fundamental concept, prevalent algorithms, and various categorizations.
Section~\ref{sec:FL-C-Literature} is dedicated to research question Q1 and offers a detailed overview of the existing literature with a focus on decentralized control and learning in control, as well as on concepts at the intersection of FL and control. 
In Section~\ref{sec:Fl-and-Control}, we delve into the expected benefits of merging those fields and discuss potential applications, which addresses research question Q2. 
Finally, Section~\ref{sec:Conclusion} gives some conclusions. 
A visualization of the paper's structure is given in Fig.~\ref{fig:organigram}.
\section{Federated Learning}\label{sec:FL}
This section provides an overview of the fundamental principles of Federated Learning (FL), laying the groundwork for addressing the research questions posed earlier. 
We first introduce the primary goal of FL and a commonly used optimization algorithm. We then categorize the various applications of FL to highlight its versatility. 
The section concludes with a discussion of the key challenges inherent to FL. 

\subsection{Basic Concepts \& Algorithms}
Federated Learning emerged as a novel research area following the pioneering work at Google, see~\cite{konevcny2016_Federated_Optimization} and ~\cite{mcmahan2017communication}.
This work introduced a distributed model training approach particularly aimed at preserving privacy.
The general idea presented is that devices - mainly mobile phones - download the current model from a central server, learn from the local, private data, and then transmit, using encrypted communication, the model updates back to the central server.
The server immediately aggregates these updates to refine the shared global model.
Importantly, all training data remains on the device, and individual updates are not stored in the cloud, see~\cite{mcmahan2017federated_blog}.
The present survey does not claim to give a comprehensive description and discussion of all existing FL algorithms, methods, and technologies in the context of FL.
% At this point, it should be noted that the present work does not encompass a complete description and discussion of all existing algorithms, methods, and technologies in the context of FL. 
However, it focuses on the connection between FL and control.
For an in-depth exploration of FL in general, the readers are referred to the literature, such as~\cite{kairouz2021advances},~\cite{yang2019federated} and~\cite{wang2021field}, and IEEE standards~\cite{IEEE_Guide_to_FL_2021, yang2021_IEE_FL_white_paper}.
 
Federated Learning encompasses a collection of directives and algorithms tailored for distributed, privacy-preserving, and communication-efficient learning. 
While this definition is informal, it can be precisely formulated as a distributed optimization problem expressed by the following mathematical representation
\begin{align} \label{eq:FL-objective}
   \theta^* \leftarrow \arg \min_\theta \mathbb{E}_{i\sim \mathcal{P}} \left[ \mathbb{E}_{x_i \sim \mathcal{D}_i} \left[ f_i(x_i, \theta) \right] \right].
\end{align}
Here, the objective is to find the optimal value $\theta^*$, which minimizes the expected value $\mathbb{E}_{i \sim \mathcal{P}}$, taken over clients $i$ sampled from the client distribution $\mathcal{P}$, of the expected value $\mathbb{E}_{x_i \sim \mathcal{D}_i}$ of the local client cost functions $f_i(x_i, \theta)$, parameterized by $\theta$ and conditioned on the data $x_i$ sampled from the local data-generating distribution $\mathcal{D}_i$.

Assuming standard regularity conditions,~\eqref{eq:FL-objective} is typically solved using gradient descent techniques, wherein the global gradient is approximated as the expected value of local gradients.
% Under standard regularity assumptions, the optimization of the FL objective in~\eqref{eq:FL-objective} can be performed using gradient descent, approximating the global gradient as the expected value of local gradients. % \textcolor{red}{Why can we do this? --> Linearity of expectation}  https://www.wikiwand.com/en/Leibniz_integral_rule#Measure_theory_statement 
%
This involves utilizing sample averages over the local data-generating distributions $\mathcal{D}_i$ to estimate the local gradients.
Subsequently, the expected value of the global gradient is computed through some form of weighted averaging of these local gradients.
Consequently, only gradient information and never any raw data $x_i$ are exchanged, facilitating the privacy-preserving and communication-efficient optimization of the global objective function.

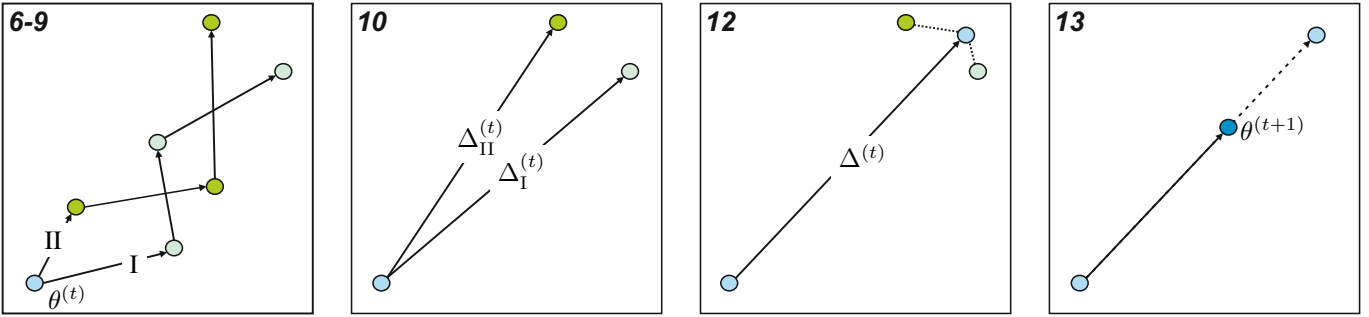
\begin{figure*}[h]
  \centering
    %\includepdf{img/gef_fed_avg.pdf}
    \begin{overpic}[width=\textwidth]{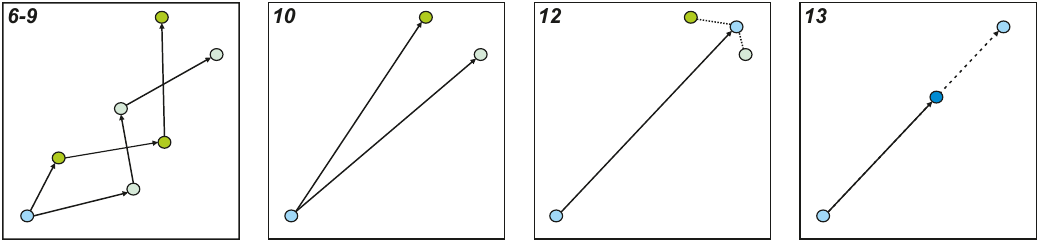}
        \put(3.6,0.6){$\theta^{(t)}$}
        \put(9,3.25){\colorbox{white}{\Romannum{1}}}
        \put(2.75, 5){\colorbox{white}{\Romannum{2}}}
        \put(36,10){\colorbox{white}{$\Delta_{\mathrm{I}}^{(t)}$}}
        \put(33,12.5){\colorbox{white}{$\Delta_{\mathrm{II}}^{(t)}$}}
        \put(61, 11){\colorbox{white}{$\Delta^{(t)}_{}$}}
        \put(91, 13){$\theta^{(t+1)}$}
        %\put(75,5){$\theta^{(t+1)} = \theta^{(t)} + \eta_s \Delta$}
    \end{overpic}
    \caption{Sketch of the optimization procedure of Generalized Federated Averaging for two clients.}
  \label{fig:gen_fed_avg}
\end{figure*}

A widely adopted FL algorithm, known as \emph{Federated Averaging}, introduced in~\cite{mcmahan2017communication}, performs the stochastic approximation of the global gradient through a strategy of \emph{partial participation} of the clients. 
This involves approximating the expected value over the client distribution $\mathcal{P}$ through a sample average over a number of clients $N$, coupled with executing \emph{local steps}, encompassing multiple iterations of local gradient steps.
The standard Federated Averaging can be extended to provide flexibility in choosing both the local and global optimization method, resulting in the \emph{Generalized Federated Averaging} algorithm, also known as \emph{FedOpt}.
This extension is detailed in~\cite{reddi2021adaptive} and presented in Algorithm~\ref{alg:generalized_fedavg}.
Examining Algorithm~\ref{alg:generalized_fedavg}, it becomes evident that three key degrees of freedom govern the optimization of the FL objective in~\eqref{eq:FL-objective}, namely
\begin{itemize}
    \item Local optimization (client-side),
    \item Global optimization (server-side), and
    \item Model aggregation.
\end{itemize}
Importantly, these components can be independently chosen, paving the way for a diverse family of algorithms capable of solving a manifold of different problems.
\RestyleAlgo{ruled}[h]
\begin{algorithm}
\caption{Generalized FedAvg (FedOpt) algorithm.}\label{alg:generalized_fedavg}
\SetAlgoLined
\textbf{Input:} Initial model $\theta^{(0)}$, ClientOpt \& ServerOpt with learning rates $\eta$, $\eta_s$\;
\For{$t = 0, 1, \ldots, T$} {
    Sample a subset $S_t$ of clients from $\mathcal{P}$\;
    \For{client $i \in S_t$ in parallel} {
        Initialize local model $\theta_i^{(t,0)} \leftarrow \theta^{(t)}$\;
        \For{$k = 0, \ldots, \tau_i - 1$}{
            Compute local gradient $g_i(\theta_i^{(t,k)})$\;
            Local update: $\theta_i^{(t,k+1)} = \textrm{ClientOpt}(\theta_i^{(t,k)}, g_i(\theta_i^{(t,k)}), \eta, t)$\;
            }
        Compute local changes $\Delta^{(t)}_i = \theta_i^{(t,\tau_i)} - \theta_i^{(t,0)}$\;
    }
    Aggregate local changes $\Delta^{(t)} = \sum_{i \in S_t} p_i \Delta^{(t)}_i / \sum_{i \in S_t} p_i$\;
    Update global model $\theta^{(t+1)} = \textrm{ServerOpt}(\theta^{(t)},\Delta^{(t)}, \eta_s, t)$;
}
\end{algorithm}
The optimization procedure of the \emph{Generalized Federated Averaging} algorithm is depicted in detail in Fig.~\ref{fig:gen_fed_avg}.
Starting from the global model $\theta^{(t)}$, steps 6-9 entail \textbf{local optimization}, where numerous local gradient steps are performed locally for clients \Romannum{1} and \Romannum{2}. 
Subsequently, in step 10, these steps are consolidated into local model updates $\Delta_{\mathrm{I}}^{(t)}$ and $\Delta_{\mathrm{II}}^{(t)}$  w.r.t. the initial model for each client.
The \textbf{model aggregation} in step 12 combines these local model updates to generate a global model adjustment $\Delta^{(t)}$, where $p_i$ is introduced to weigh the impact of each client.
Finally, step 13 illustrates the \textbf{global optimization}, wherein a single gradient step $\theta^{(t+1)} = \theta^{(t)} + \eta_s \Delta^{(t)}$ with global learning rate $\eta_s$ is executed. 
Typically, steps 4-13 are reiterated until convergence of the global model is achieved.
Apart from the proposed method for solving~\eqref{eq:FL-objective}, FL can also be formulated as a consensus optimization problem - tackled using primal-dual methods for distributed convex-optimization, see~\cite{jakovetic2020_primal-dual-method-for-distributed-opt} - or as the problem of privacy-sensitive fusion of multiple probability density functions, see~\cite{koliander2022_fusion_of_prob_density_functions}.

\subsection{Categorizations}
The FL objective in~(\ref{eq:FL-objective}) enables various applications, often categorized  by client heterogeneity, see~\cite{mitra2021_FL_client_Heteroneity}, and learning tasks. 
An alternative FL formulation, learning an implicit mixture of the global and pure local models, is presented in~\cite{hanzely2020_FL_as_mixture_of_local_and_global}. 
A comprehensive survey and classification of FL is provided in~\cite{wahab2021survey}. 
The most common distinction is between cross-device and cross-silo FL.
\textbf{Cross-device FL} embodies the original concept of collaboratively learning a shared model across a large number of devices, as detailed in~\cite{mcmahan2017communication} and~\cite{mcmahan2017federated_blog}.
Conversely, \textbf{cross-silo FL}, introduced in~\cite{kairouz2021advances}, becomes particularly relevant for smaller sets of clients.
Here, every client engages in each round of learning and maintains a state that describes the current model and its evolution.
The primary characteristics of both approaches are summarized and compared with distributed learning in Table~\ref{tab:distributed_silo_device}, see~\cite{kairouz2021advances}.
The key characteristics are emphasized in a dark gray shade, where the main differentiating features are the orchestration and client state.
\begin{table*}[h]
\caption{Comparison of distributed learning, cross-silo and cross-device FL, adapted from~\cite{kairouz2021advances}.} 
\label{tab:distributed_silo_device}
\centering
    \begin{tabular}
    {|>{\columncolor{yellow!30}}m{2cm}|>{\columncolor{black!0}}m{4.5cm}|>{\columncolor{black!0}}m{4.5cm}|>{\columncolor{black!0}}m{4.5cm}|}
    \hline
    \rowcolor{black!00}  & \textbf{Distributed learning} & \textbf{Cross-silo FL} & \textbf{Cross-device FL} \\
    \specialrule{0.2em}{0.0em}{0.0em} % Adjust thickness as needed
    \cellcolor{black!0} \textbf{Setting / Clients} & Training a model on a large dataset - clients are computing nodes in a single cluster or datacenter & Training a model on siloed data - clients are different organizations or entities & Clients are a very large number of mobile or IoT devices
    \\ 
    \hline 
    \rowcolor{black!0} \textbf{Data distribution} & Data is centrally stored and can easily be shuffled and balanced across clients - any client can read any part of the dataset & \multicolumn{2}{l|}{\parbox{9cm}{Data is generated locally and remains decentralized - each client stores its own data and cannot read the data of other clients}} \\
    \hline
    \rowcolor{black!20} \textbf{Orchestration} & \multicolumn{2}{l|}{\parbox{9cm}{All clients are almost always available}} & Only a fraction of clients is available at any one time \\
    \hline
    \cellcolor{black!0} \textbf{Scale} & Typically, $10^3$ clients & Typically, $10^0-10^2$ clients & Massively parallel, up to $10^{10}$ clients \\
    \hline
    \cellcolor{black!0} \textbf{Primary bottleneck} & Computation, as very fast communication is available in datacenters & Computation or communication & Communication \\
    \hline
    \cellcolor{black!0} \textbf{Addressability} & \multicolumn{2}{l|}{\cellcolor{black!0}\parbox{9cm}{Each client has an identity or name that allows the system to access it specifically}} & Direct indexing of clients is not possible\\
    \hline
    \rowcolor{black!20} \textbf{Client state} & \multicolumn{2}{l|}{\parbox{9cm}{Stateful – each client may participate in each round of the computation, carrying a state}} & Stateless – each client will likely only participate once or a few times \\
    \hline
    \cellcolor{black!0} \textbf{Connection} & Stable & Relatively stable & Highly unreliable \\
    \hline
    \cellcolor{black!0}\textbf{Data-partition axis} & Arbitrary across the clients & Partition is fixed — horizontal
    or vertical & Partition is fixed and horizontal \\ \hline
    \end{tabular} 
\end{table*}

\begin{figure*}
    \centering
    \begin{subfigure}[b]{0.4\textwidth}
        \centering
        \begin{overpic}[width=\textwidth]{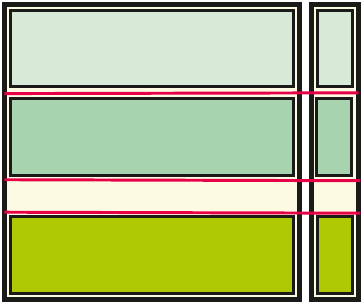}
            \put(37,85){Features}
            \put(85.5, 85){Targets}
            \put(35, 68.5){Client 1}
            \put(90.5, 68.5){1}
            \put(35, 43.75){Client 2}
            \put(90.5, 43.75){2}
            \put(35, 13){Client $N$}
            \put(90.5, 13){$N$}
            \put(45, 28){$\vdots$}
            \put(91, 28){$\vdots$}
        \end{overpic}
        \caption{Horizontal federated learning (HFL)}
        \label{fig:HFL}
    \end{subfigure}
    \hspace{0.05\textwidth} % Adjust the horizontal space between subfigures
    \begin{subfigure}[b]{0.4\textwidth}
        \centering
        \begin{overpic}[width=\textwidth]{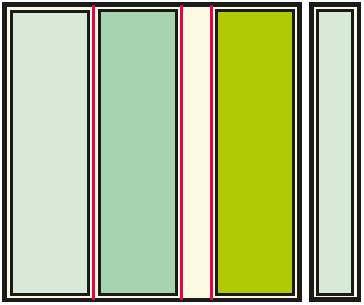}
            \put(37,85){Features}
            \put(85.5, 85){Targets}
            \put(7, 39){Client 1}
            \put(91, 39){1}
            \put(31, 39){Client 2}
            \put(61, 39){Client $N$}
            \put(51, 39){$\dots$}
        \end{overpic}
        \caption{Vertical federated learning (VFL)}
        \label{fig:VFL}
    \end{subfigure}
    \caption{Sketch of the data partition for horizontal (a) and vertical federated learning (b). Note that only one client (Client 1) provides target values for VLF. The red lines indicate the splits through the centralized data matrix.}
    \label{fig:HFL_and_VLF}
\end{figure*}

Another common distinction in FL is horizontal and vertical FL, primarily based on the data space, as introduced in~\cite{yang2019federated} and visualized in Fig.~\ref{fig:HFL_and_VLF}.
Adopting the notation from~\cite{yang2019federated}, \textbf{horizontal FL}, also known as sample-based FL, is employed when clients share the same feature space $\mathcal{X}$ and target space $\mathcal{Y}$ but differ in the sample space $\mathcal{I}$.
This can be thought of as a horizontal split through the large data matrix we would obtain in a centralized setting (clients send their data to a central server).
We formalize this as
\begin{align}
    \mathcal{X}_i = \mathcal{X}_j, \ \mathcal{Y}_i = \mathcal{Y}_j, \ \mathcal{I}_i \ne \mathcal{I}_j \ \forall \ \mathcal{D}_i, \mathcal{D}_j, i \ne j,
\end{align}
wherein $\mathcal{D}$ denotes the local data.
In horizontal FL, the objective is to address the same learning task across all clients, involving identifying the functional relationship between data samples drawn from $\mathcal{X}_i$ and $\mathcal{Y}_i$ for each client $i$.
A prominent instance of this approach can be found in mobile phones, for example, in next-word prediction, emoji suggestion, and out-of-vocabulary word discovery, as detailed in~\cite{mcmahan2017federated_blog}.
It is crucial to emphasize that the FL task within horizontal FL is effectively defined only when all clients share their true functional relationship realized through their data-generating processes. 
While minor disparities, e.g., through different noise levels, can be accommodated, more significant deviations in the functional relationship necessitate the application of advanced techniques such as clustered FL, as discussed in~\cite{sattler2020clustered, briggs2020federated, taik2022_clustered_vehicular_FL}, or federated meta-learning (FedMeta), see~\cite{liu2023federated_meta_learning}.

\textbf{Vertical FL}, referred to as feature-based FL, is employed when clients share the sample space $\mathcal{I}$ but differ in the feature space $\mathcal{X}$, as outlined in~\cite{yang2019federated}.
This scenario can be formalized as
\begin{align}
    \mathcal{X}_i \ne \mathcal{X}_j, \ \mathcal{Y}_i \ne \mathcal{Y}_j, \ \mathcal{I}_i = \mathcal{I}_j \ \forall \ \mathcal{D}_i, \mathcal{D}_j, i \ne j.
\end{align}
It can be conceptualized as instances where the large data matrix we would obtain in a centralized setting undergoes vertical splits across multiple clients; see Fig.~\ref{fig:VFL} for a graphical depiction~\cite{xia2021_vertical_FL_horizontally_partitioned_labels}. 
Vertical FL is applied in privacy-preserving regression tasks, as demonstrated in~\cite{gascon2016secure} or~\cite{hardy2017private}, as well as in healthcare~\cite{nvidia2019clara} and finance~\cite{nvidia2022finance}.
Note that the differences in target space $\mathcal{Y}$ are implicit in the task definition, as only one client (in our case client 1 as marked by the same color in Fig.~\ref{fig:VFL}) can provide targets to regress on.
It is further important to highlight that vertical FL necessitates integration with supplementary privacy-preservation methods like differential privacy, homomorphic encryption, or secure multi-party computation, further elaborated in~\cite{dwork2008differential, acar2018survey, bogdanov2008sharemind, mcmahan2017_learning_DP_recurrent_language_models}, respectively, as sharing some encrypted data is necessary.
Additionally introduced in\cite{yang2019federated} is the niche concept of Transfer FL or federated transfer learning, applied in situations at the boundary between horizontal and vertical FL.
In this context, transfer learning techniques, as delineated in~\cite{pan2009survey}, are employed in the FL task, see~\cite{liu2020_secure_FTL, saha2021_federated_transfer_learning}.
This involves learning a shared feature representation from a small common data set, which can be applied subsequently to generate predictions for samples with features from only one client. 
For further details, see~\cite{yang2019federated, razavi2022federated, liu2020_secure_FTL}.

Another significant categorization is based on the global model. 
In the prevalent approach of FL, also termed \textbf{centralized FL}, a central entity - the global server - is responsible for creating and managing the global model and steering the federation process. 
This approach offers benefits such as ease of implementation and increased control over client sampling. 
However, drawbacks include heightened vulnerabilities to system failures and adversarial attacks against a single point of failure, the global server. 
In contrast, \textbf{decentralized FL} has no central entity, as discussed in detail in~\cite{beltran2023decentralizedFL_survey, fallah2020_personalized_FL_meta-learning_appraoch}. 
Instead, some form of decentralized model aggregation occurs over a peer-to-peer network, aiming to address the downsides of centralized FL mentioned above.
Another categorization is introduced in~\cite{xu2021_edge_intelligence}.
Herein, additional servers (edge server) are introduced, allowing for near real-time responses to a smaller number of clients, leading to the concept of \textbf{edge-based FL}.
Combining the advantages of edge servers with a central server then introduces \textbf{hierarchical FL}, see~\cite{liu_2020_hierarchical_FL, zhang2020_mobile_edge_intelligence} for in-depth discussions of hierarchical FL.

\subsection{Challenges} \label{subsec:FL-challenges}
Implementing FL methods and algorithms on dedicated applications is associated with various technical challenges, discussed in detail in~\cite{khan2021federated, chellapandi2023federated, lim2020federated_survey, li2020federated}.
Despite the communication-efficient distributed learning, there is still a substantial \textbf{communication overhead} compared to local learning, with the constant exchange of model updates between the clients and a central server. 
This is particularly challenging in systems with limited bandwidth, potentially leading to delays, increased latency, and increased resource utilization, see~\cite{konevcny2016_FL_improving_communication_efficiency, luping2019_CMFL}. 
The \textbf{heterogeneity} of the clients' computational resources poses another hurdle, demanding adaptive strategies to manage collaborative learning across devices with diverse computational power, memory, and energy resources, see~\cite{wang2020_tackling_heterogenity_in_FL, mitra2021_FL_client_Heteroneity, imteaj2022_FL_resoruce_constraint_IoT, imteaj2021_survey_FL_resource-constrained_IoT, nishio2019_client_selection_FL_heterogeneous_resources}. 
The non-independent and identically distributed (\textbf{non-IID}) nature of data across clients introduces further complexities in model training, necessitating algorithms that account for variations in local data distributions, see\cite{zhao2018_FL_non-iid_data, sattler2019_robust_FL_from_non-iid_data}. 
Finally, the challenge of \textbf{model aggregation complexity} arises from the intricate process of combining asynchronous updates, addressing learning rate discrepancies, and ensuring robust mechanisms for aggregating disparate local models into a coherent global model, see~\cite{pillutla2022_robust_aggregation_for_FL}. 
Addressing these challenges is pivotal to successfully integrating FL in systems with diverse clients and data distributions.

In addition to the technical challenges, FL faces critical hurdles related to evaluation criteria, digital ethics, and incentive mechanisms. 
\textbf{Robust diagnostics} capable of identifying and eliminating updates from clients with faulty sensors or incorrect data are essential for ensuring the reliability and accuracy of the aggregated FL models, see~\cite{imteaj2020_FedAR_resource-aware_FL}. 
On the ethical front, \textbf{privacy and security} issues are paramount~\cite{woisetschläger2024federated}. 
FL's core principle of preserving user privacy while training models without exchanging raw data is a delicate balance, requiring advanced encryption techniques to aggregate model updates without compromising sensitive information. 
However, inherent security issues, including data poisoning, adversarial attacks, and the potential for reconstructing private raw data, are present challenges to FL systems, see~\cite{gosselin2022privacy}. 
Designing effective \textbf{incentive mechanisms} for FL presents a formidable challenge as participants must be motivated to contribute their computational resources and data without direct access to the immediate benefits of the global model, see\cite{zhan2021incentive, zeng2022federatedadaptivecontrol_TWC, xu2021_multiagent_FRL_incentive_mechanism}. 
This balancing act requires strategies to encourage active and sustained participation in the collaborative learning process.
Despite the mentioned challenges, a key objective of FL is to obtain strong, personalized models for individual clients, see~\cite{deng2020_adaptive_personalized_FL, jiang2019_improving_FL_personalization_via_MAML, mansour2020_3_appraoches_to_FL_personalization, wang2019_fed_personalization_evaluation}.

%%%%%%%%%%%%%%%%%%%%%%%%%%%%%%%%%%%%%%%%%%%%%%%%%%%%%%%%%%%%%%%%%%%%%%%%%%%%%%%%%%%%%%%%%%%
%%%%%%%%%%%%%%%%%%%%%%%%%%%%%%%%%%%%%%%%%%%%%%%%%%%%%%%%%%%%%%%%%%%%%%%%%%%%%%%%%%%%%%%%%%%
\section{State-of-the-Art at the Intersection of FL and Control} \label{sec:FL-C-Literature}

Building on the prior brief introduction to Federated Learning (FL), this section focuses on the state-of-the-art at the intersection of FL and control.
First, the concept of \emph{distributed control} is introduced. 
Following that, we investigate \emph{learning and control}, concentrating on system identification and controller design. 
Here, we provide an overview of the literature we deem interesting for FL. 
Finally, we provide a concrete overview of the current literature at the intersection of FL and control.
\subsection{Distributed Control}
%\begin{itemize}
%    \item What is DC, DO?: basic framework, assumptions, applications
%    \item Similarities / Synergies between DC and FL
%    \item Differences to FL
%\end{itemize}
%
Distributed Control (DC) is a subfield of control that focuses on designing and implementing control systems where the control task is distributed across multiple connected controllers.
This allows for a robust framework for managing complex and large-scale systems where centralized control would  be less effective.
Each controller in the system manages its local operations and integrates into the global control strategy through a network of communication links to other controllers, see~\cite{siljak2011_decentralized_control_complex_systems, dullerud2004_distributed_control_heterogeneous_systems, tse1983_distributed_control_linear_systems, dAndrea2003distributed_controller_design_spatially_interconnected_systems, trentesaux2009_distributed_ctrl_production_systems, baggio2021_datadriven_control_complex_networks, liu2011_controllability_complex_networks}.
Again, the scope of this section is not to provide a detailed introduction to DC, as there is excellent literature available, e.g., ~\cite{antonelli2013_interconnected_dynamic_systems, cao2012_overview_distribtued_multi-agent_coordination}, but to provide the main concepts, assumptions, and some examples. 
Similarities and synergies between DC and FL, as well as major differences, are also discussed. 
Earlier work on distributed multi-agent coordination can be found in, e.g.,~\cite{tsitsiklis1984_problems_in_decentralized_decision_making, tsitsiklis1986_distributed_deterministic_and_SGD_algorithms, murray2007_recent_reserach_in_cooperative_control}.
A further overview focusing on decentralized control is given by~\cite{bakule2008_decentralized_control_overview}.

In the context of DC, several foundational assumptions are essential.
DC is based on the \textbf{decentralization of control}, which means that each controller has enough computational power and resources to manage local control tasks autonomously.
This decentralization demands robust \textbf{connectivity} for the efficient exchange of information, guaranteeing that controllers can coordinate and synchronize operations throughout the system.
Nevertheless, the DC architecture is also designed so that the failure of individual components does not trigger a systemic failure, thereby ensuring continuous system operation despite localized disturbances.
\textbf{Scalability} is another critical assumption underlying DC. 
The system design should allow for a seamless addition or removal of nodes and modifications to the network’s configuration without significant disruptions to its overall functionality.
\textbf{Synchronization} mechanisms within DC are assumed to ensure that, despite the autonomy of individual controllers, operations are harmonized across the system, particularly where processes are interconnected~\cite{cao2012_overview_distribtued_multi-agent_coordination, zhou2020_distributed_control_networked_microgrids}. 
Moreover, the effectiveness of DC depends on the availability of accurate and timely \textbf{data from local sensors}, which each node uses to make informed decisions. 
This reliance on data assumes that all nodes have access to the necessary resources - such as computational power and energy - to perform their functions reliably.
Finally, given the distributed nature of the system, \textbf{security measures} are presumed to be robust, safeguarding data privacy against potential adversaries.
These assumptions serve as the foundation for the design of DC systems.

Established application examples for DC include smart grids, multi-agent systems, and connected autonomous
vehicles (CAVs). 
In \textbf{smart grids}, various components like renewable energy sources and consumer appliances must coordinate to distribute electricity efficiently, e.g.,~\cite{zhao2015_distributed_control_microgrids, molzahn2017_survey_distributed_opt_and_control_electric_power_systems, guo2017_distributed_control_and_opt_in_smart_grid_systems+, zhou2020_distributed_control_networked_microgrids}. 
This coordination is done using DC, where each node adjusts its operations based on local data, e.g., energy demand or supply availability, and communicates with neighboring nodes to maintain stability and efficiency.
The work~\cite{khalil1978_control_using_different_models_same_system} also provides an early example of applying ideas from DC to large-scale systems, illustrated by a power system example, whereas~\cite{kalathil2015_online_learning_for_demand_response} provides a summary of recent research.

In distributed control of \textbf{multi-agent systems}, see~\cite{lewis2013_cooperative_control_multi-agent-systems, tang2023_zeroth-order_feedback_opt_cooperative_multiagent_systems}, formation control involves multiple agents maintaining a predefined spatial arrangement while moving, see, e.g.,~\cite{wang2022_mean-field_game_control_swarm_formation}.
An early example of formation control of unmanned aerial vehicles is provided by \cite{mutambara1994_distributed_robot_control}.
Each robot adjusts its position based on the positions of its neighbors rather than central instructions, see~\cite{stipanovic2004_decentralized_control_unmaneed_aerial_vehicles}.
In~\cite{demir2012_cooperative_control_multi-agent-systems_event-based_communication}, control performance and communication effort are compared in multi-agent systems.

Furthermore, the \textbf{platooning of connected automated vehicles} (CAVs) is projected to significantly alter road transportation by enhancing traffic efficiency and decreasing fuel consumption, see~\cite{li2017_dynamical_modeling_distributed_control_CAVs, li2017_distributed_platoon_control}.
In~\cite{zheng2016_distributed_MPC_vehicle_platoons}, a distributed model predictive control algorithm is proposed for the vehicle platooning problem.
The rapid development of vehicle-to-vehicle communications, see~\cite{noor2022_6G_for_V2X} for a detailed discussion of vehicle-to-everything (V2X) communication, also encourages using distributed control techniques. 
The work~\cite{wang2019_survey_cooperative_long_motion_control_CAVs} provides a detailed survey on cooperative longitudinal motion control for multiple CAVs.

Further examples are given in~\cite{demir2011_decomposition_approach_decentralized_and_distributed_control}, wherein decentralized, distributed and centralized control systems are compared based on the objective of improved system performance of a multi-zone furnace.  
The authors of~\cite{gambuzza2020_distributed_control_of_multiconsensus} discuss the problem of achieving multi-consensus in a linear multi-agent system using distributed controllers.
The distributed control of nonlinear interconnected systems is studied in~\cite{xu2022_distributed_observer_and_control_affine_nonlinear_systems}.
State observers are also investigated in the setting of distributed control.
For example, \cite{wang2017_distributed_observer_LTI, mitra2018_distributed_observer_for_LTI, park2016_design_distributed_LTI_observer} discuss the important issue of distributed observer design for LTI systems, whereas~\cite{kim2016_distributed_Luenberger} focuses on the design of distributed Luenberger observers.
In~\cite{khan2011_stability_optimality_distributed_KF}, the distributed Kalman filter is discussed.

DC and FL share several foundational principles based on their decentralized nature. 
Both methodologies emphasize decentralization, where decision-making for DC and learning for FL are distributed across agents rather than centralized in a single entity. 
This structure reduces the risk associated with single points of failure, increasing the system's robustness.
Both DC and FL rely heavily on local processing. 
Agents in these systems handle data locally, reducing the need to transport vast amounts of data across the network. 
This not only saves bandwidth but also improves privacy by limiting the accessibility of critical information. 
Scalability is another shared feature; DC and FL are designed to handle an increasing number of agents easily. 
This scalability means that growing the network or integrating more agents does not affect the overall system performance, allowing the system to remain manageable and operationally efficient as it grows.
These shared characteristics underline the adaptability and efficiency of both DC and FL in managing complex, distributed tasks. 
Whether in controlling physical processes or in data processing for machine-learning, both paradigms leverage decentralization, local processing, and coordinated communication to meet the respective goals effectively.

While Distributed Control (DC) and Federated Learning (FL) share some fundamental principles, their differences are significant, reflecting different aims, applications, and operating approaches.

\textbf{Purpose and Application:} DC is commonly used in engineering systems to control physical processes and devices, such as grids, fleet robotics, and infrastructure management. 
Its main goal is the direct control of physical entities to ensure operational efficiency and reliability - fulfilling the global control task.
In contrast, FL is a machine-learning technique that enables the training of models across multiple decentralized devices and is particularly useful in data-driven applications where privacy and bandwidth constraints are critical.

\textbf{Data Handling:} Data handling also underscores a fundamental difference between the DC and FL. 
DC is concerned with real-time control, whereas FL uses data to train machine-learning models without real-time considerations.

\textbf{Algorithmic Focus:} From an algorithmic perspective, DC concentrates on algorithms that ensure stability, control, and optimization of system dynamics, reflecting its direct interaction with physical systems. 
On the other hand, FL’s algorithms are geared towards learning and inference, aiming to optimize accuracy, reduce model bias, and, importantly, enhance data privacy through its decentralized approach.

\textbf{Privacy Concerns:} In terms of privacy, while DC may handle sensitive information, especially in critical infrastructure settings, its primary concerns revolve around operational security and system reliability. 
FL explicitly addresses privacy concerns, as it is designed to minimize data exposure by ensuring that only model updates are communicated rather than raw data.

In summary, while DC and FL operate on distributed principles, they apply these principles to fundamentally different tasks - control and optimization in real systems for DC and privacy-preserving collaborative learning in FL. 
Their methodologies reflect their respective goals: immediate and direct control of physical environments versus incremental and privacy-preserving improvement of predictive models.

\subsection{Learning and Control with Focus on FL}
%
% This section briefly introduces \emph{Learning and Control} (LC), focussing on concepts and applications similar to FL.
%As LC has the potential to fill textbooks, we do not intent to give a deep introduction, but only a brief exploration of the topics we consider relevant in relation to FL.
%We cite relevant literature and encourage the reader to emerge into this amazing field at the intersection of artificial intelligence and control theory. 
This section briefly introduces \emph{Learning and Control} (LC), focusing on concepts similar to Federated Learning (FL). 
While LC deserves extensive coverage, we provide a concise overview of relevant topics and refer readers to the key literature to explore this exciting intersection of artificial intelligence and control theory.
Within the field of LC, we see two main categories that are promising for FL, namely
\begin{itemize}
    \item System identification
    \item Controller design
\end{itemize}
The review of data-driven control~\cite{hou2013_survey_DDC_MBC} offers insights into the challenges of model-based control (MBC) theory, the significance of data-driven control (DDC) methods, the state-of-the-art, their classifications, and the relationship between model-based and data-driven control.

\subsubsection{System Identification}
Standard textbooks describe various ways to tackle the problem of system identification, such as~\cite{Ljung.1991, chiuso2019systemID_ML_perspective, nelles2020_nonlinear_dynamic_System_identification}.
In the following, we discuss selected works in this field that apply ideas from the learning literature to system identification problem.

For the identification of stable LTI systems,~\cite{umenberger2018_max_likelihood_id_stable_LDS} proposes a maximum likelihood routine based on the Expectation Maximization (EM) algorithm with latent states or disturbances and Lagrangian relaxation.
The identification of unstable linear systems is discussed in~\cite{faradonbeh2018_finite_time_id_unstable_lin_sys}, where finite-time bounds on the error of the Least Squares (LS) estimate of the dynamic matrix are derived for a large class of heavy-tailed noise distributions. 
The work~\cite{sarkar2019_near_optimal_finite_time_id_LDS} presents a novel statistical analysis of the LS estimator for LTI systems with stable, marginally stable, and explosive dynamics.  
Learning autonomous linear dynamical systems from a single trajectory or rollout is thematized in~\cite{simchowitz2018_learning_without_mixing, oymak2019non-asymp-ID-LTI-single-trajectory}, whereas~\cite{xin2022identifying_dynamics_similar_systems} and~\cite{tu2022_learning_from_many_trajectories} discuss identification based on multiple trajectories.
In~\cite{hazan2017_learning_LTI_spectral_filtering, hazan2018_spectral_filtering_general_linear_DS} spectral filtering is introduced to learn the impulse response function for latent-state linear dynamical systems.
When working with nonlinear systems of the form $\dot x = f(x, u)$, a common practice is linearizing around a reference point $\{x_r, u_r\}$. 
The work~\cite{xin2023learning_linModels_nonlinSystems} follows this idea and uses regularized LS to obtain the linearized dynamics. 

An overview of nonlinear dynamic system identification techniques is given in~\cite{schoukens2019_nonlinear_system_id}.
A perspective based on kernel-methods and Bayesian statistics, including support vector machines, Gaussian regression, and reproducing kernel Hilbert spaces (RKHS), is proposed in~\cite{chiuso2019systemID_ML_perspective} and further emphasized in~\cite{pillonetto2022regularizedSysID}.
A kernel specifically tailored for Port-Hamiltonian systems is presented in~\cite{Beckers2022phsKernel} and preserves the passive nature within this system class. 
Gaussian mixture models are applied to encode the robot's motion as a first-order autonomous nonlinear ODE, see~\cite{khansari2011_learn_stable_nonlinear_dynsys_with_GMM}.
In~\cite{kaiser2018_SINDY_with_MPC}, the combination of SINDY - sparse identification of nonlinear dynamics, see~\cite{brunton2016_sparse_ID_nonlinear_dynamics_with_control} - and MPC is proposed to enhance the control performance.
A modern perspective based on deep neural networks is elaborated in~\cite{pillonetto2023_deep_NN_sys_id} and~\cite{legaard2023_NN_for_dyn_systems}.
Using neural networks in system identification was already proposed in the 1990s~\cite{sjoberg1994_neural_net_sys_id, narendra1990_identification_and_control_using_NN, hunt1992neural}.
The authors of~\cite{achille2018_separation_principle_for_control_and_DL} formulate the system identification task based on high-dimensional uncertain measurements, e.g., videos, as a neural network-based approximation of the posterior of the control loss.  
Deep Learning was also applied to learn representations of the Koopman operator and its eigenfunctions from data, as seen in~\cite{lusch2018_deep_learning_universal_linear_embeddings_nonlin_dynamics}.
Another line of work focuses on \emph{Neural State Space models}, wherein neural networks are utilized to learn state space representations.
Early works trace back to~\cite{zamarreno1998_state_space_neural_networks}, where recurrent neural networks are used for nonlinear system identification.
Recent works utilize autoencoders~\cite{masti2021_learning_nonlin_SSM_using_autoencoders, beintema2021_nonlinear_statespace_identifcatin_deep_encoder_networks}, genetic algorithms~\cite{skomski2021_automating_PINN_SSM}, or meta-learning~\cite{chakrabarty2022meta-learning-bayesopt-closed-loop-performance, lew2022_safe_active_dynamics_learning_and_control} for learning a nonlinear state space representation.
The problem of learning long sequence dependencies is tackled via structured state space models in~\cite{gu2021_efficiently_model_long_sequences_structure_state_spaces, gu2021_efficiently_model_poster, gu2023mamba}.
Learning state representations can be interpreted as a generalization of system identification, as it comprises learning forward and inverse models.
An excellent review summarizing state representation learning formalism, as well as the learning objectives and building blocks, are given in~\cite{lesort2018_state_representation_learning}.

Table~\ref{tab:lin_nonlin_sysid_potential_FL} summarizes recent advances in learning-based identification for linear and nonlinear systems, which we consider promising for FL applications.
\begin{table*}[h!t]
\caption{Recent literature regarding linear and nonlinear system identification with potential for FL.}
\label{tab:lin_nonlin_sysid_potential_FL}
\centering
\begin{threeparttable}
    \begin{tabular}{|p{0.1\linewidth}|p{0.35\linewidth}|p{0.1\linewidth}|p{0.1\linewidth}|p{0.1\linewidth}|p{0.1\linewidth}|} \hline
    \textbf{Source} & \textbf{Description} & \textbf{Dynamics} & \textbf{Techniques} & \textbf{FL potential} & \textbf{Input} \\     
    \specialrule{0.2em}{0.0em}{0.0em}
    \cite{chang2022distributed_online_sysID_LTI} & Introduces distributed stochastic gradient descent (SGD) with reversed experience replay for distributed online system identification of identical LTI systems & LTI & SGD & High & None \\ \hline
    \cite{toso2023clustered_SysID_LTI} & Clustered system identification based on mean squared error (MSE) criterion & LTI & SGD; Cluster estimation & High  & Gaussian  \\ \hline
    \cite{chen2023multiTask_SysID_LTI} & System identification inspired by multi-task learning to estimate the dynamics of linear time-invariant systems jointly by leveraging structural similarities across the systems via regularized LS & LTI & regularized LS; Proximal gradient method & High & Constant \\ \hline \
    \cite{formentin2021_control_oriented_regularization_in_sysID} & Propose \emph{Control-oriented regularization} for LTI system identification using control specifications as Bayesian prior & LTI & Bayesian perspective; regularized LS & Medium & Linear \\ \hline 
    \cite{zheng2020_nonasymptotic_id_LTI_multiple_trajectories} & Provide finite-time analysis for learning Markov parameters of LTI systems applying an ordinary LS estimator with multiple rollouts covering both stable and unstable systems & part. observable LTI (open-loop stable or unstable) & LS; Markov parameters & High & Gaussian \\ \hline
    \cite{simchowitz2019_learning_LTI_semiparametric_LS} & Prefiltered Least Squares algorithm that provably estimates the dynamics of partially-observed linear systems & part. observable LTI & prefiltered LS & Medium & Gaussian \\ \hline 
    \cite{xin2023learning_dynamics_similar_systems} & Leverage data from auxiliary (similar) systems  & LTI & weighted LS & Medium & Gaussian  \\ \hline 
    \cite{modi2024joint_learning_LTI_dynamics} & Jointly estimating transition matrices of multiple, related systems & LTI & SGD; Basis functions & High & None \\ \hline 
    \cite{khosravi2019_controller_tuning_bayes_opt} & Automated tuning method for controller with safety constraints & Linear & Bayes opt.; Nonlin. regression & High & PI-control \\ \hline
    \cite{bao2020_identification_of_State_space_LPV_using_ANN} & Simultaneously estimate states and explore structural dependencies between estimated dynamics & Linear parameter varying (LPV) & multiple neural networks & Medium & Necessary\tnote{1} \\ \hline
    \specialrule{0.2em}{0.0em}{0.0em} % Adjust thickness as needed \\
    \cite{helwa2017_multi_robot_transfer_learning} & Multi-robot transfer learning for SISO systems & Nonlinear & Transfer learning & Medium & Input-output linearization \\ \hline
    \cite{zamarreno1998_state_space_neural_networks}, \cite{forgione2021_dynonet} & Neural state space models & Nonlinear & Neural network & High & Possible\tnote{2} \\ \hline
    \cite{masti2021_learning_nonlin_SSM_using_autoencoders, beintema2021_nonlinear_statespace_identifcatin_deep_encoder_networks} & Neural state space models via autoencoder & Nonlinear & neural network; Autoencoder & Medium & Necessary\tnote{1} \\ \hline
    \cite{lew2022_safe_active_dynamics_learning_and_control} & Safe learning and control based on an online uncertainty-aware meta-learned dynamics model & Nonlinear & NN; Last layer adaptation & High& Necessary\tnote{1} \\ \hline 
    \cite{arcari2021_bayesian_multi_task_learning_using_trigonometric_functions} & Bayesian multi-task learning model using trigonometric basis functions to identify errors in the dynamics & Nonlinear & Basis functions; Max. Likelihood; Kalman filtering & Medium & MPC \\ \hline
    \cite{mckinnon2021_meta_learning_forward_and_inverse_models_for_SMPC} & Stochastic MPC based on a learned input-feature model combined with (online) Bayesian linear regression and online model selection to leverage multiple input-feature models & Nonlinear &  Bayesian; NN; & High & MPC \\ \hline
    \cite{zhan2022_calibrating_building_simulation_models_multi_source_data} &  Probabilistic Deep Learning to meta-learn building models using multi-source datasets & Nonlinear & NN; meta-learning; & High & None \\ \hline
    \end{tabular}
    \begin{tablenotes}
     \item[1] Inputs are necessary for system identification, but not specified in the source.
     \item[2] Method also applicable for autonomous systems. 
    \end{tablenotes}
\end{threeparttable}
\end{table*}

\subsubsection{Controller Design}
When given a system description, through first-principles modeling or data-driven system identification, we typically want to control the system to behave in a desired way. 
For this, a controller which determines the system's input is designed.
The main distinction here is between controllers without feedback (feedforward) and with feedback, wherein a feedback controller uses measurements of the system's output and closes the control loop.
Learning controllers using neural networks have a rich history, exemplified in works like~\cite{hunt1991_NN_for_IMC, hunt1992neural, narendra1990_identification_and_control_using_NN, grant1989_NN_pole_balancing}.
Moreover, data-driven control (DDC) is thematized in~\cite{dePersis2019_formulas_for_data_driven_control, dePersis2023_learning_controller_for_nonlin_from_data}, with a special focus on control design for nonlinear systems.
A distinction can be made between direct DDC and indirect DDC, where system identification and control are performed sequentially. 
In~\cite{dorfler2022_bridging_direct_indirect_DDC}, the connection between indirect and direct DDC approaches is discussed. 
They formulate these approaches using behavioral systems theory and parametric mathematical programs and bridge them through a multi-criteria formulation, trading off system identification and control objectives. 
The study reveals that direct DDC can be derived as a convex relaxation of the indirect approach, with regularization accounting for implicit identification steps.
Direct and indirect predictive control is the main topic of~\cite{krishnan2021_direct_vs_indirect_predictive_control}.
In this comparative study based on stochastic LTI systems, two distinct non-asymptotic regimes in control performance can be distinguished for direct and indirect predictive control.

There is also much work on optimal control, especially for unknown, or partially known systems.
In~\cite{ouyang2017_learning_based_control_unknown_lin_sys_with_thompson_sampling}, a Thompson sampling-based learning algorithm is used to learn the dynamics, which are subsequently used for Linear Quadratic Regulator (LQR) control design.
They show robustness to time-varying parameters of the controlled stochastic LTI system.
Controlling an LTI system with known noisy dynamics and adversarial quadratic loss is tackled using semi-definite relaxation in~\cite{cohen2018online_LQR} and~\cite{cohen2019_learning_LQR_efficient}, leading to strongly stable policies.
LQR control for unknown linear systems is further investigated in~\cite{dean2018_regret_bounds_robust_adaptice_control_LQR, dean2019_safely_learning_to_control_constraint_LQR, dean2020_sample_complexity_of_LQR, ferizbegovic2019_learning_robust_LQR_LTI}.
The problem of adversarial changing convex cost functions with known linear dynamics is tackled in~\cite{agarwal2019_logarithmic_regreti_for_online_control} and~\cite{agarwal2019_online_learning_adverserial_disturbances} using a Disturbance-Action Controller (DAC) given by $u_t = K x_t + \sum_{i=1}^H M_{i-1} w_{t-i}$ combined with online convex optimization. 
The papers~\cite{hazan2020_nonstochastic_control_problem, simchowitz2020_making_nonstochastic_control_easy, simchowitz2020_improper_learning_for_nonstochastic_control, chen2021_black-box-control-LTI} focus on controlling unknown linear dynamical systems subjected to non-stochastic, adversarial perturbations.
The class of Gradient Perturbation Controllers (GPC) is introduced, combining a stabilizing linear controller $K$ with a DAC parametrized by the matrices $M_i, i = 1, \dots, H$ for some horizon $H$ and disturbance $w_i$. 
The general class of convex optimization control policies (COCPs), including standard applications like LQR, approximate dynamic programming, and model predictive control is discussed in~\cite{agrawal2020_learning_convex_OCP}.
They propose updating the control parameters based on the projected stochastic gradients of performance metrics (cost functions) instead of the standard way of tuning by hand, or by grid search.
The work of~\cite{zheng2021LQG_unknown_dynamics} considers Linear Quadratic Gaussian (LQG) problems with unknown dynamics. 
They leverage Input-Output Parameterization (IOP), see~\cite{furieri2019_input_output_parameterization}, for robust controller synthesis based on a convex parameterization.
Similar works~\cite{lale2021_adaptive_control_in_LQG} and~\cite{lale2020_logarithmic_regret_boud_PO_LDS} focus on the adaptive aspect of the problem and introduce a control algorithm combining system identification and adaptive control for LGQ systems.

In adaptive control, early work uses neural-network-based adaptive controllers for trajectory tracking of robotic manipulators, see~\cite{sun1999_NN_adaptive_controller_robotic_manipulator_with_observer, ciliz1997_online_learning_control_manipulators_based_on_ANNs, jin1993_stable_NN_control_for_manipulators}.
Radial basis function networks were used even earlier for adaptive control, see~\cite{sanner1991_gaussian_networks_direct_adaptive_control}.
Recent work on model reference adaptive control (MRAC) based on Gaussian Processes is given in~\cite{joshi2018_adaptive_control_GP-MRAC}, whereas~\cite{joshi2019_deep_MRAC} studies MRAC based on deep neural networks.
In~\cite{hanover2021_adaptive_NMPC_for_quadrotors}, an adaptive nonlinear MPC is designed, so that model uncertainties are learned online and immediately compensated for. 
Adaptive control for high-dimensional systems is always challenging.
This problem is tackled in~\cite{boffi2021_learning_stability_certificates_from_data}, wherein a non-parametric adaptive controller is proposed that scales to high-dimensional systems by learning the unknown dynamics in a reproducing kernel Hilbert space (RKHS) leveraging random Fourier features.
The work~\cite{nguyen2008_local_GPR_online_learning} introduces local Gaussian process regression as a method that achieves high accuracy in online learning and real-time prediction, applied to inverse dynamics model learning for model-based control of two 7-DoF robot arms.

Table~\ref{tab:ctrl_design_potential_FL} summarizes the recent advances in learning-based controller design, which we consider promising for FL applications.

\begin{table*}[h!t]
    \centering
    \caption{Recent literature regarding controller design with potential for FL.}
    \begin{tabular}{|c|p{0.4\linewidth}|p{0.1\linewidth}|p{0.1\linewidth}|p{0.1\linewidth}|}
    \hline
    \textbf{Source} & \textbf{Description} & \textbf{System} & \textbf{Techniques} & \textbf{FL potential}\\     
    \specialrule{0.2em}{0.0em}{0.0em}
    \cite{harrison2018_control_via_meta_learning_dynamics, harrison2019_ADAPT_zero_shot_policy_transfer_stochastic_DS} & Controlling an unknown system with varying latent parameters, aiming to learn approximate models for both the dynamics and the prior over latent parameters from observed trajectory data & Nonlinear & NN; meta-learning & High \\ \hline
    \cite{richards2021_adaptive_control_meta_learning, richards2023_control_oriented_meta_learning} & Control-oriented approach to learning parametric adaptive controllers through offline meta-learning from past trajectory data & Nonlinear & NN; meta-learning; & Medium \\ \hline
    \cite{muthirayan2022_meta_learning_online_control_LTI} &  Meta-learning online control algorithm that learns across a sequence of similar control tasks & Linear & Projected gradient descent & Medium \\ \hline
    \cite{li2023_data_based_stab_in_linear_systems} & Transferring knowledge from source system to design a stabilizing controller for target system & Linear & Transfer Learning & High \\ \hline
    \cite{berberich2022_combining_prior_knowledge_and_data_robust_control_design} & Combining prior knowledge and measured data for learning-based robust controller design, leading to stability and performance guarantees for closed-loop systems & Linear & Linear fractional transformations & Medium \\ \hline 
    \cite{shi2021_OMAC} & Online multi-task learning approach for adaptive control with environment-dependent dynamics & Nonlinear & NN; Last-layer adaptation & High \\ \hline
    \cite{zhang2023multi_task_IL_LTI} & Multi-task imitation learning via shared representations for linear dynamical systems & Linear & Multi-task imitation learning & High \\ \hline
    \cite{sun2021_online_learning_unknown_dynamics_model_based_locomotion} & Learning a time-varying, locally linear residual model of the dynamics to compensate for prediction errors of the controller's design model & Linear residual model & Ridge regression; Replay buffer & Medium \\ \hline
    \cite{rajeswaran2016_EPopt} & Combine simulated source domains and adversarial training to learn robust policies & Nonlinear & Policy gradient; NN & Medium \\ \hline
    \cite{yu2020_learning_fast_adaptation_with_MSO} & Present a learning algorithm termed \emph{Meta Strategy Optimization} that learns a latent strategy space suitable for fast adaptation in training & Nonlinear & Meta learning; NN & Medium \\ \hline
    \cite{devin2017_learning_modular_NN_policies_multitask_multirobot_transfer} &  Decomposing neural network policies into task-specific and robot-specific modules, enabling transfer learning across different robots and tasks with minimal additional training & Nonlinear & NN; multi-task \& transfer learning; & High \\ \hline
    \cite{toso2024_meta-learning_LQR_policy-gradient_MAML} & LQR in a multi-task setting, with the MAML-LQR approach producing stabilizing controllers close to task-specific optimal controllers & Nonlinear & Policy gradient \& NN; meta-learning & High \\ \hline 
    \end{tabular}
    \label{tab:ctrl_design_potential_FL}
\end{table*}

\subsection{Existing Literature for FL and Control}
Currently, existing works that combine Federated Learning (FL) methodologies with control theory can be found in four major areas:
\begin{itemize}
    \item system identification, 
    \item controller design,  
    \item federated reinforcement learning, and
    \item control-inspired aggregation.
\end{itemize}
Table~\ref{tab:FL-C-literature} summarizes the primary outcomes of these studies. 
This section refers to the first research question, Q1, in Section~\ref{sec:Intro}.
\begin{table*}[htbp]
    \centering
    \caption{Summary of the literature connecting FL and Control.} 
    \label{tab:FL-C-literature}
    \begin{tabular}{|c|p{0.7\linewidth}|}
    \hline
    \textbf{Topic} & \textbf{Summary and Literature References} \\
    \specialrule{0.2em}{0.0em}{0.0em} % Adjust thickness as needed
    \textbf{System Identification} &
    Recent research explores collaborative learning of linear systems in system identification, using FedSysID to balance performance and system heterogeneity, see~\cite{wang2023fedsysid, chen2023multiTask_SysID_LTI}.
    A distributed Recursive Least Squares (RLS) algorithm, discussed in~\cite{azzollini2023robustRLS}, addresses robust estimation in networked environments, aligning with FL principles. \\
    \hline
    \textbf{Controller Design} &
    Practical applications of Federated LQR concepts have demonstrated their scalability and efficiency, marking significant strides in both linear and nonlinear systems~\cite{ren2020federatedLQR, wang2023modelfreefederatedLQR}. In~\cite{zeng2021federatedadaptive_CDC}, the integration of neural networks into an adaptive PID-control system, leveraging FL to overcome limitations posed by limited local training data, showcases the potential of this approach. In~\cite{wang2023_fleet_policy_learning}, FL is harnessed to learn static feedback controllers for a fleet of robots, while~~\cite{shiri2020_communication-efficient_UAV_online_path_control_with_FL} explores the use of FL in swarm control of drones.
    Personalized FL is applied for learning in robotic control applications~\cite{nakanoya2021personalized}. \\
    \hline
    \textbf{Reinforcement Learning} &
    Federated Learning mitigates reinforcement learning's sample inefficiency and accommodates heterogeneous sensor data, see~\cite{lim2020federated, liu2019lifelong, liu2020federated, liang2022federated}. \\
    \hline
    \textbf{Control-inspired Aggregation} &
    Novel research integrates control theory principles into FL by enhancing global model aggregation with integral and derivative terms, akin to PID control frameworks, see~\cite{machler2021fedcostwavg, mansour2022fedcontrol, machler2023fedpidavg, gao2023fedadt}. \\
    \hline
    \end{tabular}
\end{table*}

In \textbf{system identification}, a strong focus has been put on collaboratively learning linear, time-invariant (LTI) system dynamics from diverse observations across multiple clients under privacy considerations, as exemplified in~\cite{wang2023fedsysid} and \cite{chen2023multiTask_SysID_LTI}. 
In~\cite{wang2023fedsysid}, the \emph{FedSysID} algorithm is introduced, addressing Least Squares (LS) system identification for multiple linear clients with similar dynamics, showing improved convergence compared to individual learning.
In~\cite{chen2023multiTask_SysID_LTI}, the focus lies on leveraging similarities among multiple systems to accurately estimate LTI dynamics using LS, although privacy considerations were not specifically addressed.
Another notable contribution, detailed in~\cite{azzollini2023robustRLS}, introduces a distributed Recursive Least Squares (RLS) algorithm tailored to robust estimation in networked environments. 
In this scenario, each client measures samples linked by a common, yet unknown, linear regression model. 
This algorithm can be interpreted within the framework of FL, as knowledge is exchanged between clients by transmitting and aggregating covariance matrices of an RLS algorithm. 
Pioneering work in adaptive networks\footnote{Adaptive networks consist of a collection of nodes with learning abilities. 
The nodes interact with each other on a local level and diffuse information across the network to solve inference and optimization tasks decentralized.}, a field not directly linked but conceptually related to FL, comprehensively examines recent strides in adaptation, learning, and optimization across networked systems, as surveyed in~\cite{baggio2021_datadriven_control_complex_networks} or~\cite{sayed2014adaptive}.
These surveys offer valuable insights for comparing diverse network topologies and facilitate an assessment of adaptive networks' performance in contrast to centralized implementations.

\textbf{Controller design} has a line of work on federated LQR concepts, e.g.,~\cite{ren2020federatedLQR} and \cite{wang2023modelfreefederatedLQR}. 
In~\cite{ren2020federatedLQR}, the distributed LQR tracking problem is studied in a setting of clients sharing linear but unknown dynamics and tracking different targets. 
The proposed model-free federated zero-order policy gradient algorithm capitalizes on a shared LQR matrix across clients, demonstrating a linear speedup in the number of clients over communication-free, local alternatives, even when clients have a heterogeneous component in their objective, namely their tracking target. 
Simulation results demonstrate the algorithm's effectiveness in linear and nonlinear system settings, showcasing its scalability.
In~\cite{wang2023modelfreefederatedLQR}, the focus is on the model-free federated LQR problem involving multiple clients with distinct and unknown yet similar LTI dynamics collaborating to minimize an average quadratic cost function while maintaining data privacy. 
The proposed FL approach, named \emph{FedLQR}, allows clients to periodically communicate with a central server to train policies, addressing questions related to the stability of the learned common policy, its proximity to individual clients' optimal policies, and the speed of learning with data from all clients.
In~\cite{wang2023_fleet_policy_learning}, fleet-level learning of static feedback controllers from distributed and potentially heterogeneous robotic data is discussed.
They proposed the \emph{FLEET-MERGE} algorithm, which efficiently merges neural network-based controllers by accounting for permutation invariance in recurrent neural networks without centralizing the data.

An additional line of work delves into the integration of neural networks within adaptive PID-control systems for connected autonomous vehicles (CAVs), as exemplified in~\cite{zeng2022federatedadaptivecontrol_TWC} and~\cite{zeng2021federatedadaptive_CDC}. 
Control parameters are dynamically adjusted to enable navigation under varying traffic and road conditions, which is facilitated by a neural network-based auto-tuning unit learning the system behavior, as detailed in~\cite{nie2018longitudinal}. 
The autonomous vehicle's local training data, restricted by onboard memory, limits the controller's adaptability to specific traffic scenarios, potentially compromising safety. 
To overcome this constraint, the proposed approach advocates for implementing FL.
In this collaborative process, CAVs collectively learn the neural network-based auto-tuning units facilitated by a wireless base station functioning as a global server. 
This FL mechanism empowers CAVs to enhance neural network auto-tuning units for local controller adjustments, effectively tackling the issue of limited local training data and broadening the controller's applicability across diverse traffic patterns.
In~\cite{nakanoya2021personalized}, personalized FL is applied for learning trajectory forecasting models in robotic control applications, wherein personalization is achieved by adjusting learning rates based on parameter variances. 
An excellent example of the combination of FL and control is given in~\cite{shiri2020_communication-efficient_UAV_online_path_control_with_FL}.
Here, the swarm control of a large number of drones is considered, combining mean-field game theory, Federated Learning, and Lyapunov stability analysis.
Numerical examples show the efficacy of the proposed approach.

The intricate connections between control and \textbf{reinforcement learning} (RL) arise from their common pursuit of effective decision-making and optimization.
Control theory contributes well-established principles for system regulation and trajectory tracking, complemented by reinforcement learning's adaptive techniques designed for acquiring optimal behaviors in dynamic and uncertain settings; see, e.g.,~\cite{bertsekas2019reinforcement, levine2020offline,vidyasagar2023tutorial} for detailed reviews of reinforcement learning.
The growth in computational power and the development of deep neural networks have significantly enhanced the capabilities of reinforcement learning algorithms.
This technological growth has propelled the field into new frontiers, enabling more sophisticated learning and decision-making processes.
Reinforcement learning algorithms, however, are confronted with the challenge of sample inefficiency, often necessitating a substantial number of interactions with the environment to acquire effective policies. 
This inefficiency poses a significant drawback, particularly in real-world scenarios where data collection can be resource-intensive and time-consuming.
Federated Learning  techniques offer a promising avenue to mitigate these challenges by facilitating collaborative learning among multiple agents (clients) to derive optimal policies, see~\cite{lim2020federated}.
The authors propose a federated RL scheme based on the actor-critic Proximal Policy Optimization (PPO) algorithm to learn a classical nonlinear control problem: the upswing of a pendulum.
They clearly show the effectiveness of the proposed FL solution by significantly reducing the convergence time, despite the slightly different dynamics of the individual devices, namely three \emph{Quanser QUBETM-Servo 2}, see~\cite{Quanser}. 
In~\cite{khodadadian2022_FRL_linear_speedup_markovian_sampling}, a federated reinforcement learning setup shows a linear convergence speedup concerning the number of agents. 
Additionally, FL's capability to accommodate heterogeneous real-life or artificially generated sensor data further broadens its applicability and effectiveness in diverse and complex environments, as exemplified in~\cite{liu2019lifelong, liu2020federated, liang2022federated, kumar2017_federated_control_multi-agent_deep_RL}.

The discussion so far predominantly centers around applying FL techniques within specific control areas, namely system identification, controller design, and reinforcement learning.
However, the landscape looks different when examining \textbf{control-inspired aggregation}, representing a shift in viewpoint.
In this context, concepts derived from control are employed to enhance FL algorithms. 
A noteworthy research path involves augmenting the global model aggregation process by incorporating integral and derivative terms akin to conventional PID-control frameworks. 
This approach, exemplified in works such as~\cite{machler2021fedcostwavg, mansour2022fedcontrol, machler2023fedpidavg, gao2023fedadt}, represents a unique integration of control theory principles to optimize the FL process.
Another work, conceptually similar to FL, is given by~\cite{kalathil2015_online_learning_for_demand_response}.
Here, they tackle the problem of consumer scheduling under incomplete information.

\section{FL and Control} \label{sec:Fl-and-Control}
In light of the literature review conducted in the previous section, this section explores various scenarios in which Federated Learning (FL) concepts can be integrated into control theory, encompassing system identification and control, and knowledge transfer in multi-agent systems. 
Furthermore, FL's adaptive learning capabilities are discussed for dynamic systems, highlighting its suitability for evolving environments and changing operating conditions. 
This section refers to the second research question, Q2, in Section~\ref{sec:Intro} and presents FL as a promising framework for advancing various facets of control.

\subsection{Methodical Benefits of Combining FL and Control}
In control, the integration of FL principles offers a promising avenue for advancing system adaptability via communication-efficient, collaborative learning with preserved privacy as well as enhanced generalization and robustness.
Federated Learning's \textbf{adaptive learning} capabilities find resonance in control applications dealing with dynamic systems. 
Clients can continuously update their knowledge using a system model or controller parameterization based on changing environmental conditions. 
This makes FL suitable for control scenarios where system dynamics or shared external influences evolve.
The decentralized approach inherent in FL aligns seamlessly with distributed control systems. 
This synergy enables multiple clients engaged in control tasks to collaboratively update their control policies or global model representation based on local observations, fostering a coordinated and efficient system behavior, as demonstrated in~\cite{lim2020federated} and~\cite{qi2021federated}.
The emphasis on \textbf{communication efficiency} is particularly beneficial for control systems with many participants or low bandwidths. 
Clients engaged in collaborative learning tasks can exchange minimal data while acquiring control policies, mitigating communication challenges, and enhancing overall system efficiency.
\textbf{Transfer learning} proves valuable in control domains, mirroring FL's practice of training models on one task and adapting them to another, see~\cite{xiang2021_fast_CRDNN_on-site_training_mobile_construction_machines}.
This concept facilitates transferring and fine-tuning of learned control policies from one system to a similar yet distinct one, minimizing the necessity for extensive  and time-consuming retuning  and retraining efforts.
\textbf{Privacy-preserving control} emerges as a critical application, leveraging FL's emphasis on local model training without raw data sharing. 
In scenarios where data privacy is paramount, such as personal data in healthcare, critical infrastructure, or strictly confidential production data, clients can collaboratively learn control policies without disclosing sensitive information, thus aligning with privacy requirements.
By integrating robust control techniques into FL algorithms, further guarantees of system performance under uncertainties or adversarial conditions may be established, bolstering the reliability and generalization of learned models in real-world control applications.

In summary, including FL principles, encompassing adaptability, generalization, communication efficiency, decentralization, and privacy preservation, holds significant potential for advancing various facets of control. 
This integration contributes to developing more efficient, flexible, secure, and responsive control systems.

\subsection{Expected Applications of Combining FL and Control}

\begin{table*}[ht]
\caption{Overview of Expected Applications.}
\label{tab:applications}
\centering
\begin{threeparttable}
    \begin{tabular}{|p{4cm}|p{4cm}|p{4cm}|p{4cm}|}
    \hline
    \textbf{Application} & \textbf{Core Idea} & \textbf{Benefits} & \textbf{Potential Downsides} \\
    \specialrule{0.2em}{0.0em}{0.0em} % Adjust thickness as needed
    Parametric federated SysID & Inferring parameters of the structurally known dynamics of a system based on local input-output data from multiple clients & Improved convergence rate over single client learning; persistent excitation of global model; \emph{federated dynamics} & FL costs\tnote{1}~; useful only for similar systems; high level of system knowledge required  \\
    \hline
    Non-parametric federated SysID & Collaboratively learn non-parametric dynamics based on local input-output data from multiple clients & Improved convergence; increasing generalizability; \emph{federated dynamics} &  FL costs\tnote{1}~; useful only for similar systems \\
    \specialrule{0.2em}{0.0em}{0.0em} % Adjust thickness as needed
    Indirect Adaptive Control & Learning \emph{federated dynamics} model and use local or global model as robust control design model & Improved control performance and robustness under uncertainties in dynamic environments & FL costs\tnote{1}~; similar control requirements necessary; \\
    \hline
    Direct Adaptive Control & Learn a control law on local data and aggregate at global server & Improved generalization; improved initial model for local adaptive control; & FL costs\tnote{1}~; application to similar systems only; potential stability issues \\
    \hline
    Advanced server-side optimization & Utilizes cloud computing power on global server & Server-side update/training of neural-network-based control laws for low power hardware, e.g., Learning-based NMPC approximation & FL costs\tnote{1}~; powerful server necessary \\
    \specialrule{0.2em}{0.0em}{0.0em} % Adjust thickness as needed
    Multi-agent decision-making & Explicit distinction between decision-oriented and support-oriented agents & Improves privacy-preserving decision-making and coordination among multiple agents& FL costs\tnote{1}~;  \\
    \hline
    Sensor fusion \& shared representation learning & Combine different sensor modalities to extract a richer environment representation & Enhances situational awareness and perception capabilities, e.g., merging visual, depth, and semantic maps & FL costs\tnote{1}~; scalability issues \\
    \hline
    Transfer learning & Merges real-life and simulator data for knowledge transfer & Facilitates safer exploration \& knowledge transfer & FL costs\tnote{1}~;  \\
    \hline
    \end{tabular}
     \begin{tablenotes}
     \item[1] See Section~\ref{subsec:FL-challenges} for a detailed discussion of FL costs.
   \end{tablenotes}
\end{threeparttable}%
\end{table*}
There are many ways to include concepts from FL in control.
Table~\ref{tab:applications} summarizes the introduced applications.

\subsubsection{System Identification}
Foremost, Federated Learning presents a promising avenue for enhancing \textbf{system identification} (SysID) within the control framework. 
Typically, system identification is separated into parametric SysID and non-parametric SysID.
In parametric SysID, first principles and process knowledge are used to develop parametric models, whose physics-based parameters are calibrated based on input-output data. 
In non-parametric SysID, black-box models, e.g., highly-capable function approximation algorithms like neural networks, are employed to learn the system dynamics.
Federated Learning extends this paradigm by enabling multiple clients to collaboratively learn from decentralized data sources while preserving data privacy, leading to federated SysID. 

In this context, \textbf{parametric federated SysID} allows individual clients to leverage their input-output data to calibrate local models, which are subsequently aggregated to refine a global model representing the parametric system dynamics, subsequently termed \emph{federated dynamics}.
This ranges from sharing single parameter information to learning the complete dynamics of some LTI or nonlinear system.
Sharing information through the global model not only copes with parameter scattering caused by minor, not modeled system differences, e.g., component-aging effects, but also allows for some form of persistent excitation of the global model; see~\cite{narendra1987persistent} for further information regarding persistent excitation.
Furthermore, the scalability and generalizability of the obtained \emph{federated dynamics} are enhanced, and concerns regarding data privacy and security are addressed. 
An improved convergence rate can be obtained compared to individual client learning.
For instance, smart pneumatic or hydraulic valves can come equipped with series models that capture the nonlinear input-output relationships in a parametric form.
During operation, these valves encounter distinct application-specific operating conditions and device-specific variations like wear and tear, often not accounted for during end-of-line calibration.
The valves can adapt to such scenarios by embedding a local learning algorithm.
These adaptations can also be fed back to the manufacturer in a privacy-preserving manner to enhance the overall series model.
As a note on the downside, challenges inherent to FL, as discussed in Section~\ref{subsec:FL-challenges}, are also inherent for parametric SysID when combined with FL.
A high level of system knowledge is also required to formulate the parametric models.
Parametric federated SysID can only be applied when the participating clients and systems follow similar dynamics. 
A possible remedy here is the application of clustered FL techniques, wherein the clients are clustered into similar groups sharing similar dynamics,  see~\cite{sattler2020clustered, briggs2020federated, taik2022_clustered_vehicular_FL}.
This cluster assignment is either based on a priori known criteria or obtained through applying clustering techniques on the client model parameters. 

In non-parametric SysID, the system dynamics are typically learned from input-output data using sophisticated function approximation algorithms like neural networks or Gaussian Processes.
\textbf{Non-parametric federated SysID} enables the learned client models to update a global model collaboratively, representing the \emph{federated dynamics}.
This again fosters more robust and generalizable models of dynamical systems and facilitates a more comprehensive exploration of the systems dynamics.
Furthermore, a faster convergence typical to FL is obtained if the client dynamics are similar. 
Additionally, the computational power of the global server can be exploited for further model refinement during the global model update. 
This can enforce knowledge-based constraints similar to the approach underlying the physics-informed neural networks (PINNs), which seems intractable for embedded hardware; see~\cite{thuerey2021pbdl} for more information on physics-based Deep Learning or~\cite{Nghiem2023pinn} for physics-informed Machine Learning.
FL can also support the development of confidence models, which estimate the reliability of learned \emph{federated dynamics} based on contextual features. 
These models provide valuable insights into the confidence level associated with predictions, enabling better decision-making in dynamic environments where uncertainty is prevalent.
Integrating FL with confidence modeling enhances the trustworthiness of purely data-driven dynamical system models, contributing to more effective control and decision-making processes.
On the downside are again the challenges inherent to FL; see Section~\ref{subsec:FL-challenges}.
Moreover, that non-parametric SysID can only be successfully applied to similar systems. 
Again, clustered FL techniques, as discussed in~\cite{sattler2020clustered, briggs2020federated, taik2022_clustered_vehicular_FL}, may be helpful here. 

\subsubsection{Controller Design}
Furthermore, FL techniques can also be applied to adaptive control problems, where indirect and direct adaptive control methods are explored. 
% indirect AC
\textbf{Indirect adaptive control} involves learning a system model or its parameters from distributed data and deriving a controller based on the acquired model, as exemplified by the neural-network-based PID auto-tuning~\cite{nie2018longitudinal}.
Federated Learning can obtain a \emph{federated dynamics} model, serving as a feedback controller design model for individual clients.
The local controller can then be designed using the \emph{federated dynamics} model or a locally adapted version to ensure a high control performance and robustness under local uncertainties.
On the downside are again the challenges inherent to FL; see Section~\ref{subsec:FL-challenges}.
Also, similar control requirements on the client level are necessary to successfully apply indirect adaptive control combined with FL.

% direct AC
On the other hand, \textbf{direct adaptive control} entails learning the client controller directly from its datasets.
Federated Learning enables combining these local client controllers in a robust, privacy-preserving way. 
The resulting global controller model ensures robustness and generalizability in various dynamic settings.
For instance, hydraulic excavators can use data-driven servo-compensation models to improve path or trajectory tracking for tasks like leveling or grading, see~\cite{weigand2021hybrid}.
This servo-compensation may be equipped with a local learning algorithm, adapting individual machines to various environmental conditions, e.g., the temperature influencing the viscosity of the hydraulic oil. 
This information can be shared by applying FL to obtain a global servo-compensation model, leading to a general improvement in machine performance. 
Conversely, we again face challenges intrinsic to FL, as elaborated in Section~\ref{subsec:FL-challenges}.
Additionally, there are potential stability issues related to the global model. 

% comp. power of the global server for complex, nonlinear control algorithms
Another way to use FL is to employ the global server's computational power, termed \textbf{advanced server-side optimization}, typically much larger than the client devices, for further advanced control tasks, such as optimal control or nonlinear model predictive control (NMPC).  
Especially for the latter, the hardware requirements for a real-time implementation are still a significant cost factor. 
In~\cite{Hertneck2018approxMpc, AbuAli.2022, Karg.2021}, this hurdle is overcome by optimizing the corresponding operating range already in the development phase and subsequently training a neural network (NN) to reproduce the solution in a computationally efficient way. 
Federated Learning enables an extension of this framework towards online adaptations, as the central server can perform the computationally intensive optimization utilizing the \emph{federated dynamics} and then distribute the solution to the clients, e.g., in the form of neural networks.
The global server must have suitable computational hardware to perform this optimization.
Also, challenges related to FL and federated SysID, as already mentioned before, must be considered.

\subsubsection{Multi-Agent Decision-Making}
Progressing from system identification over controller design to knowledge transfer, FL holds promise for advancing \textbf{multi-agent decision-making} scenarios by facilitating collaborative decision-making among agents, including both decision-oriented and support-oriented agents, see~\cite{qi2021federated}. 
In such settings, FL enables agents to learn and share knowledge while preserving the privacy of their local data. 
Decision-oriented agents responsible for making critical decisions can benefit from FL by leveraging insights from decentralized data sources to improve decision-making accuracy and robustness. 
Support-oriented agents providing auxiliary functions such as data preprocessing or information aggregation can enhance their capabilities using FL by collectively learning from distributed datasets. 
Additionally, FL fosters the development of decentralized coordination mechanisms, allowing agents to coordinate their actions efficiently without centralized control. 
By employing FL techniques, multi-agent systems can adapt and evolve, leveraging the collective intelligence of diverse agents to achieve more effective and resilient models, exemplified in~\cite{liu2019lifelong} and termed \emph{Lifelong Federated Reinforcement Learning}.
For instance, decision-oriented mobile working machines in unstructured environments may be assisted by support-oriented drones equipped with visual sensors to solve tasks like handling cargo or material transport collaboratively. 

Furthermore, FL offers significant potential in multi-agent systems for learning a \textbf{shared representation} of the environment, see~\cite{liang2020_think_locally_act_globally}. 
By collaboratively training models on decentralized data sources, FL enables agents to collectively understand the environment, leading to more coherent decision-making and coordination among agents. 
This becomes particularly relevant in settings like construction sites where numerous stakeholders interact, making privacy-preserving learning and adaptation of shared environmental representations through FL methods of utmost interest.

Moreover, FL can enhance \textbf{sensor fusion} capabilities within multi-agent systems by leveraging data from diverse sensor modalities distributed across agents, see~\cite{liu2020federated}, wherein visual data (RGB and depth maps) and semantic segmentation data are combined within an FL setting. 
This enables the extraction of richer and more comprehensive information about the environment, improving situational awareness and decision-making accuracy. 

Additionally, FL facilitates \textbf{transfer learning} and \textbf{multi-task learning} (see~\cite{smith2017_federated_multi-task_learning}) in control by allowing agents to transfer knowledge learned from one environment to another.
This is exemplified in~\cite{liang2022federated}, where sophisticated simulators are combined with real-life LIDAR data to navigate autonomous model cars in changing indoor environments with obstacles.
Similarly, the work~\cite{li2019_FL_SLAM} introduces an FL architecture for cooperative simultaneous localization and mapping (SLAM) in cloud robotic systems.
This adaptability is particularly valuable in dynamic environments where conditions may change over time. 
By leveraging transfer learning techniques with FL, multi-agent systems can effectively adapt to new scenarios and improve overall performance and robustness.
Inherent to the presented applications regarding knowledge transfer, namely multi-agent planning, shared representation learning, sensor fusion, and transfer learning, are the already introduced challenges of FL, see Section~\ref{subsec:FL-challenges}.

\section{Conclusion} \label{sec:Conclusion}
This paper presents an overview of state-of-the-art methods and ideas for combining Federated Learning (FL) and control.
A detailed literature review reveals a scarcity of research at the intersection of FL and (nonlinear) control.
Building on the advancements in distributed control and learning, combining FL and control can improve the system's adaptability and privacy preservation by allowing for decentralized controller updates, privacy-preserving control, and adaptive learning, thereby improving control efficiency, security, and responsiveness.

Practically, FL enriches control methodologies by enhancing system identification, improving dynamical system modeling, and advancing control techniques.
Additionally, it facilitates knowledge transfer in multi-agent decision-making via environment representation learning, sensor fusion, and transfer learning, fostering adaptive and resilient control systems. 
Apart from individual scientific activities in the synergetic combination of FL and control, several further developments are required.
The evident mismatch in the main objectives between FL and control - FL attempts to achieve privacy-preserving global model learning, whereas control usually concentrates on localized optimal performance - may be the main reason for this open research potential, see Section~\ref{sec:FL-C-Literature}. 
Tables~\ref{tab:lin_nonlin_sysid_potential_FL} and~\ref{tab:ctrl_design_potential_FL}, combined with insights presented in Section~\ref{sec:Fl-and-Control}, provide important guidance and ideas for applying FL to system identification and control.
Up to the author's knowledge, this work marks the first comprehensive survey of FL and control, laying the groundwork for further research to combine FL and control for collaborative control applications.
In essence, FL offers a valuable additional feature set that, when integrated with control, can drive advancements across various domains. 
However, substantial further research is required to realize the ideas outlined in this work.

\bibliographystyle{IEEEtran}
\bibliography{IEEEabrv, main}

\end{document}

%% file: prefix.tex
\IEEEoverridecommandlockouts
\usepackage{cite}
\usepackage{amsmath,amssymb,amsfonts}
\usepackage{graphicx}
\usepackage{subcaption}
\usepackage{textcomp}
\usepackage{xcolor}
\usepackage{tcolorbox}
\usepackage{booktabs}
\usepackage{makecell} % for thicker lines (\Xhline{})
\usepackage{overpic}

\usepackage[linesnumbered]{algorithm2e}% http://ctan.org/pkg/algorithm2e
\makeatletter
\newcommand{\nosemic}{\renewcommand{\@endalgocfline}{\relax}}% Drop semi-colon ;
\newcommand{\dosemic}{\renewcommand{\@endalgocfline}{\algocf@endline}}% Reinstate semi-colon ;
\let\oldnl\nl% Store \nl in \oldnl
\newcommand{\nonl}{\renewcommand{\nl}{\let\nl\oldnl}}% Remove line number for one line
\usepackage{romannum}

\usepackage{float}  % load float package first!
\usepackage{hyperref} % let hyperref patch the float package stuff
\usepackage{enumitem}% http://ctan.org/pkg/enumitem
\usepackage{threeparttable}

\usepackage{pgfplots}
\usepackage{pdfpages}
\pgfplotsset{compat=1.18}
\usepackage{array}
\usepackage{colortbl}
\usepackage{hhline}
\definecolor{headercolor}{RGB}{255,204,153}
\definecolor{settingcolor}{RGB}{255,255,204}
\definecolor{distributedcolor}{RGB}{255,229,204}
\definecolor{crosssilocolor}{RGB}{255,204,153}
\definecolor{crossdevicecolor}{RGB}{204,229,255}
\definecolor{commoncolor}{RGB}{255,204,204}

\def\BibTeX{{\rm B\kern-.05em{\sc i\kern-.025em b}\kern-.08em
    T\kern-.1667em\lower.7ex\hbox{E}\kern-.125emX}}